\ifdefined\pdfoutput\pdfoutput=1\fi

\documentclass[10pt, a4paper]{article}

\usepackage[T2A,T1]{fontenc}      
\usepackage[utf8]{inputenc}       
\usepackage[russian,british]{babel}  

\usepackage[final]{lrec-coling2024}

\usepackage{graphicx}   
\usepackage{latexsym}   
\usepackage{microtype}  

\DeclareFontSubstitution{T2A}{cmr}{m}{n}

\title{A Data-Driven Approach to Idiomaticity Based on Experts' Criteria in Theoretical Linguistics}

\name{Elena Mikhalkova, Anastasiya Vishnyakova, Anastasiya Drozdova,\\
Polina Gavin, Aleksander Zhmykhov, Timofey Protasov}

\address{
Tyumen State University, Volodarskogo, 6, Tyumen, Russia\\
Aston University, Aston St, Birmingham B4 7ET, Great Britain\\
{\tt evrog2009@gmail.com, a.y.vishnyakova@utmn.ru, a.o.drozdova@utmn.ru}\\
{\tt 160232923@aston.ac.uk, linkyspie@gmail.com, t.a.protasov@utmn.ru}}

\abstract{The article observes data analysis of 286 multi-word expressions (MWEs) based on 16 lexical, grammatical and other criteria described in theoretical books and papers on the notion of idiomaticity. MWEs were collected from the same theoretical sources, and a set of experts in linguistics annotated them with these categories. The distribution of categories shows that there are no absolutely idiomatic expressions. Lexical criteria seem to be the most influential; grammatical criteria are bound to certain conditions; presence of obsolete words and grammar influence ability of an MWE to be replaced with one word.}

\begin{document}
\maketitleabstract

\section{Introduction}

Multi-word expressions (MWEs), groups of lexemes occurring in a text and more complex linguistically, than just a free word-group, can confuse automatic text processing in many ways. Even the terminology surrounding them is quite extensive and lacks commonly accepted definitions, hence, often relying on an approach to their treatment. What unites these approaches is understanding that lexemes in MWEs cannot be treated fully as their equals in free word-groups.

Works, summing up recent approaches to MWE processing, are published every now and then~\citep{pearce-2002-comparative,piao2005comparing,constant2017multiword,ashok2019comparing}. And, probably, the main world event in this topic is the annual ACL-affiliated workshop on MWEs. In recent years, we observe a trend to evaluate methods for particular languages, e.g. LREC Workshop Towards a Shared Task for Multiword Expressions (MWE 2008) addressed the issue in English, German, Czech, Estonian and other languages~\citep{gregoire2008proceedings}. Later appeared Arabic~\citep{attia2010automatic}, Russian~\citep{tutubalina2015clustering}, Polish~\citep{chrzaszcz-2016-extraction}, Spanish and Basque~\citep{inurrieta-etal-2017-rule}, Irish~\citep{walsh-etal-2019-ilfhocail}, Bulgarian and Romanian~\citep{barbu-mititelu-etal-2019-hear}, Serbian~\citep{stankovic-etal-2020-multi} and many others. Currently, there is also a trend to discuss translation and alignment~\citep{lam-etal-2015-phrase,fisas-etal-2020-collfren,han-etal-2020-alphamwe}. The last LREC workshop on MWEs~\citep{mwe-2022-multiword} addressed heuristic and machine-learning approaches to their detection, tool-kits for annotation, low-resource corpora, figurative language, etc. However, discussions about the nature of MWEs remain. In our study, we join these discussions and try to oversee one side of it that has not been granted due attention.

In our study, we describe theoretical and modern applied approaches to classification of collocations paying more attention to what can be called ``data-driven'' approaches. Second, we propose a model of 16 linguistic criteria that were derived from theoretical works by linguists. The model encompasses lexical, semantic, grammatical and pragmatic criteria. Third, we take MWEs from the same works and label them with these criteria (whether a related feature is manifested in an MWE or not). Why we use these particular method of collecting MWEs is that we are of the opinion that these examples, suggested by theorists, are a gold standard demonstrating typical features of idiomaticity. Fourth, we group the criteria into four sets and vectorize our corpus so that it can be modelled as a group of points in a 3D cube. Finally, we observe clusters of these points and conclude about how they are grouped in the multidimensional vector space and what it shows about the nature of idiomaticity.

\section{Approaches to Classification of Collocations}
\label{sec:theory}

{\bf Statistical criteria of ``MWEhood''.} \citet{baldwin-2010} underline that MWEs allow to use a comparatively brief lexicon to create nuances of meaning. The lack of freedom in MWEs, or {\em vice versa} strength of connection, is usually referred to as {\em idiomaticity} -- ``markedness or deviation from the basic properties of the component lexemes''~(Ibid.). Idiomaticity shows in ``lexical, syntactic, semantic, pragmatic, and/or statistical levels''~(Ibid.).

Statistical methods focus on inferring idiomaticity from co-occurrence of lexemes inside a certain word-group and in free contexts. Among the most common statistical measures used in this task are {\em mutual information}~\citep{church1990word}, {\em likelihood ratio tests}~\citep{dunning1993accurate}, {\em cost criteria}~\citep{kita1994comparative}. \citet{pecina2008machine} enlists 55 ``lexical association measures used for ranking MWE candidates''. The output of these methods is a number that evaluates the strength of idiomaticity. Based on it, MWEs are arranged (ranked), but are hardly classified. {\em Vice versa} the process usually leads to understanding what type of MWE is better derived with the method. It would not be wrong to state that there is no universal criterion to MWE extraction, and statistical methods are applied to existing classifications.

The term {\em collocation}, often used in papers describing statistical methods, can be considered a synonym to {\em MWE}, although \citet{baldwin-2010} put it that collocations are statistically idiomatic MWEs. In fact, we tend to observe that statistical methods extracting this of that type of MWE are designed without limitations. Hence, any MWE can look statistically significant with a proper measure. Another way to disambiguate between an MWE and collocation is that some collocations lack {\em non-compositionality} -- they are {\em compositional}, their meaning is easily extracted from lexemes composing them. E.g. {\em Many thanks!} is a statistically significant proper way of being thankful, but its meaning is clear from the words composing it. The (non-)compositionality cannot be statistically inferred from word frequencies. 

In some literature, MWEs are synonymous to {\em multi-word units} (MWUs), ``lexical items that go beyond single word items''~\citep{shin2019mwu}, and {\em words-with-spaces}, ``idiosyncratic interpretations that cross word boundaries (or spaces)''~\citep{sag2002multiword}. In our paper, we intentionally do not make any particular distinction between MWEs, collocations, MWUs and words-with-spaces, considering them to be manifestations of the same phenomenon -- idiomaticity.
 
\begin{figure}[t]
  \centering
  \includegraphics[width=0.9\linewidth]{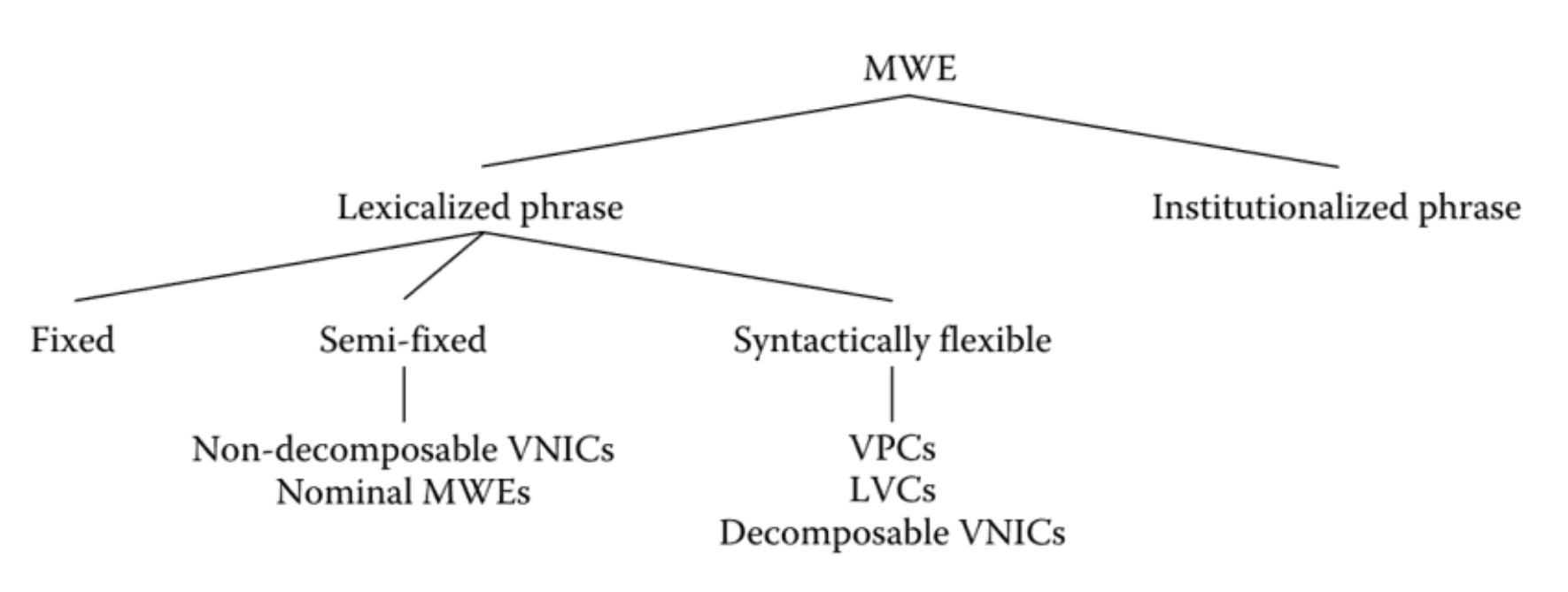}
  \caption{Classification of MWEs by (Baldwin and Kim, 2010). VNIC - verb-noun idiomatic combination; VPC - verb-particle construction; LVC - light-verb construction.}
  \label{figure_baldwin}
\end{figure}

\begin{figure}[t]
  \centering
  \includegraphics[width=0.8\linewidth]{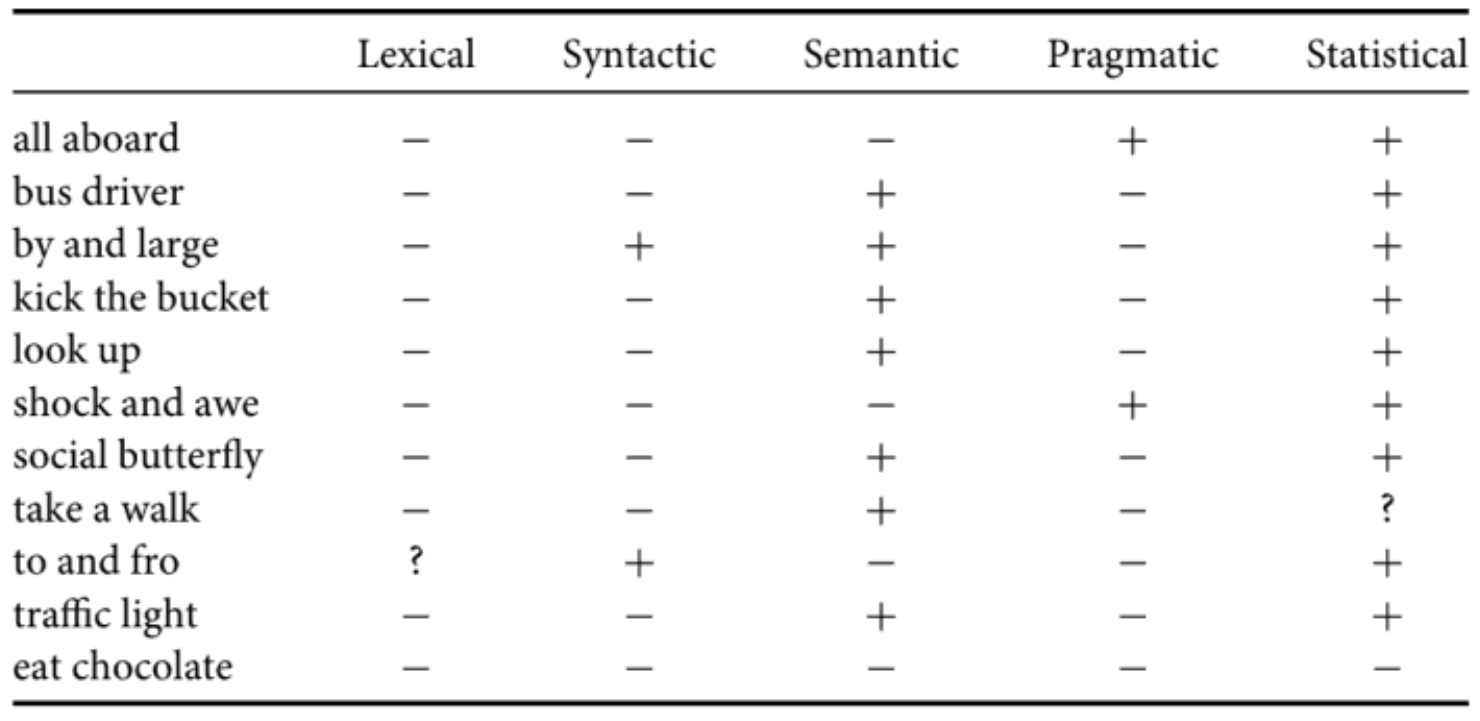}
  \caption{Classification of MWEs in terms of their idiomaticity, by (Baldwin and Kim, 2010).}
  \label{table_baldwin}
\end{figure}

{\bf Expert classifications.} \citet{baldwin-2010} suggest a classification that first splits MWEs into two main classes (see fig.~\ref{figure_baldwin}): institutionalized phrases are collocations proper (statistically common phrases such as ``Many thanks!'') and lexicalized phrases show idiomaticity to a certain degree and are marked by different features (e.g. decomposable / non-decomposable). We believe that such an approach to classification, although it is supported by examples, cannot be called data-driven; rather, it is an expert view of a complex phenomenon. Also, the bottom level of the obtained hierarchy (VNIC, nominal, VPC, LVC) is based on parts-of-speech analysis of English collocations, hence, binding idiomaticity to one particular linguistic feature. However, in Table 12.2 from the same chapter \citet{baldwin-2010} approach classification of MWEs from another perspective: they enlist properties of MWEs and annotate several examples with these properties, acquiring a matrix of feature distribution from which they conclude about the probability of ``MWEhood'', see fig.~\ref{table_baldwin}. In our opinion, this approach could be called data-driven, as classes are inferred from annotations. But the examples were few and were chosen so as to demonstrate several cases referring to pre-designed classes.

PARSEME~\footnote{\url{https://typo.uni-konstanz.de/parseme/index.php}}, a large European project that started in 2014 and focused on annotation of MWEs, split them into the following classes: 

\begin{itemize}
    \item Nominal MWEs
    \begin{itemize}
        \item Multiword named entities
        \item NN compounds
        \item Other nominal MWEs 
    \end{itemize}
    \item Verbal MWEs
    \begin{itemize}
        \item Phrasal verbs
        \item Light verb constructions
        \item VP idioms
        \item Other verbal MWEs
    \end{itemize}
    \item Prepositional MWEs
    \item Adjectival MWEs
    \item MWEs of other categories
    \item Proverbs
\end{itemize}

Resembling classification in figure~\ref{figure_baldwin}, this project lays weight on part-of-speech properties of the headword in a phrase and splits the set into distinct subgroups. We tend to believe that such an approach was organizational: it is important to split jobs in large projects. Some other possible drawbacks in it are that it has to group non-standard examples into ``other'' and that elements in the resulting hierarchy are not of the same level, theoretically. E.g., from the point of view of theoretical linguistics, named entities are represented by proper nouns (if we disregard anaphora) as opposed to all common nouns -- at the same time, Noun+Noun compounds are a subgroup of common nouns.

Another approach, yet leading to abandoning all classifications, is found in~\citep{schneider2014comprehensive}. The authors aimed to annotate a corpus for DiMSUM (Ibid.), a SemEval task for detecting minimal semantic units and their meanings. They collected a set of classes of idioms in English, that totaled 15 as illustrated in their article. Beside some of the classes, mentioned by~\citet{baldwin-2010}, it included named entities, compound words ({\em motion picture}), support and phrasal verbs ({\em make decision, cry foul}), coordinated phrases ({\em cut and dry}), phatic phrases ({\em You're welcome!}), proverbs ({\em To each his own}), etc. Annotators in this project only marked what they considered to just be an MWE. It is of interest that the authors did not find any particular POS pattern that could be associated with a collocation type: ``Categorizing MWEs by their coarse POS tag sequence, we find only 8 of these patterns that occur more than 100 times''~\citep{schneider2014comprehensive}.

In CoNLL-U Format from the Universal Dependencies project~\footnote{\url{https://universaldependencies.org/format.html\#conll-u-format}}, multi-word tokens, simply, ``are indexed with integer ranges''.

{\bf Expert criteria.} Another way to build a classification theoretically is to enlist various linguistic criteria according to which something can be defined as an MWE and terminologically label gold examples demonstrating these criteria. Such is an approach by~\citet{vinogradov1977ob}, who adapted a classification by the Swiss scholar Charles Bally and singled out two types of MWEs -- a less and a more idiomatic:

\begin{itemize}
    \item combinations ({\em to conclude an agreement})
    \item fusions
    \begin{itemize}
        \item with archaic words -- {\em to eke out}
        \item with archaic grammatical forms -- {\em hither and thither}
        \item that were changed so that they do not resemble lexemes from which they were composed -- {\em lo and behold} ({\em lo} from {\em look})
        \item complete loss of initial meaning -- {\em caught red-handed} (initially meant catching someone who hunted an animal they were not allowed to hunt)
    \end{itemize}
\end{itemize}
Outside the scope of his classification, \citet{vinogradov1977ob} placed terminological groups and named entities.

Without building a classification, \citet{manning1999foundations} describe three criteria that characterize collocations: non-compositionality (discussed above), non-substitutability (lexemes cannot be substituted with synonyms), non-modifiability (lexemes cannot change grammatically). And again several types are mentioned separately: light verbs, verb-particle constructions, proper nouns, terminological expressions.

\citet{cowie1996phraseological} suggest the following criteria: 

\begin{itemize}
    \item familiarity to speakers
    \item ability to be stored in memory as ready-made units
    \item limited and arbitrary variability
    \item opaque semantics  
\end{itemize} 

\citet{tarasevitch1991soviet} makes use of her own list:

\begin{itemize}
    \item stability of use
    \item structural separateness
    \item complexity of meaning
    \item being not built on the generative pattern of free word-groups
\end{itemize} 

\citet{melchuk1960terms} considers that idiomaticity influences translation of a phrase or its parts. In a more idiomatic and stable expression it is hard to find an exact match to every lexeme and the whole phrase is easier to translate with a singe word. \citet{baldwin-2010} also mention pragmatic idiomaticity (being associated with a certain situation), proverbiality (describing a situation of social interest), prosody, but we will leave these criteria outside the scope of our research as they require to go beyond the study of a written text.

In this paragraph, we might have overlooked some criteria, but as far as we know in other works approximately the same criteria repeat.

{\bf Which approach can be called data-driven?} \citet{amin2021data}, who call their approach data-driven in the title of their paper, design metrics that help to infer some of the mentioned above criteria for n-grams in a text. A similar scheme is traced in \citep{rossyaykin2019measure} who use a set of statistical, context and distributional measures to infer MWEs from a corpus. Another clustering method, based only on association measure, is found in~\citep{tutubalina2015clustering}. \citet{nissim2013modeling} introduce variation patterns as an alternative to association measure. \citet{wahl2018multi}, who call their approach ``bottom-up'', introduce the MERGE algorithm, again as an alternative to association measure (the project is developed by~\citet{gries2022multi}). Summing up, data-driven are projects in which MWEs are represented as vectors in multi-dimensional space and analysed statistically. Often these vectors are visualised in diagrams to see whether there are clusters that attract more MWEs.

\section{Experiment Setup}
\label{sec:experset}

The intuition lying behind our approach is that the theoretical linguistic expertise about the phenomenon of idiomaticity makes it look like a single unity that can be split into sectors -- classes, or types of MWEs. However, seeing it as an umbrella term that unites phenomena of different linguistic nature~\footnote{E.g., as mentioned, lexical: named entities versus noun compounds; grammatical: diversity of POS-patterns.} leads us to a hypothesis that we can describe this heterogeneity and outline these phenomena as clusters in a multi-dimensional space, based on vectorization of annotated examples. To visualize these clusters, we propose a data-driven approach. We believe that we do not need a large collection for such a task if we have a gold standard that manifests the main features of idiomaticity. Further, in our experiment we propose: a. a set of criteria (features of idiomaticity), b. the gold standard dataset of MWEs, c. expert annotation, and d. a method for clustering visualisation.

{\bf Criteria.} The linguistic criteria that we took from theoretical literature, described earlier, total 15. We do not claim that this list is full. The chosen criteria were picked up so as to be able to give a precise instruction on how to annotate an MWE. E.g. {\em subordinate word cannot be replaced with a synonym} is checked with an attempt to change a subordinate word in an MWE in several possible contexts, e.g. \selectlanguage{russian}``Я набрал белых грибов.'' \selectlanguage{british}({\em I've picked some penny buns.}) does not allow such a change. An example criterion that we did not include is ``Lexemes (partially) lose independent meaning'' as it is the essence of the idiomaticity and, hence, we view it as the dependent (target) variable. Another left out criterion is POS pattern: as mentioned, its importance has not been determined statistically. 

The criteria were re-formulated so as to demonstrate idiomaticity, which means that 0 denotes its lack and 1 -- its presence. An annotated example is a vector of zeros and ones, e.g. the vector for \selectlanguage{russian}``белый гриб''\selectlanguage{british} (En. {\em penny bun}) is (1, 1, 1, 1, 1, 0, 0, 0, 0, 0, 0, 1, 0, 1, 1, 1), see Table~\ref{tab16props}; its vector sum is 9. The higher is the vector sum of an annotated MWE, the more idiomatic an MWE is.

We will later describe that we visualize the scores as a 3D cube with resulting vectors. For this purpose, we grouped the mentioned 16 criteria into four categories: each category is an axis of the cube and the fourth category would be demonstrated with the color. Grouping was arbitrary, based on the levels of language ({\bf lexical} and {\bf grammatical}) and our analysis of which criteria depend on each other. E.g. if a subordinate word cannot be replaced with a synonym, then it would probably be hard to translate an MWU word by word. The analysis resulted in singling out two groups: {\bf obsolescence} -- features, related to a long use, e.g. obsolete and unique words~\footnote{To be called archaic, obsolete, a word needs other contexts. But it is unique and remains only in this particular MWE.} and {\bf replacement} -- ability of an MWE to ``shrink'' into a shorter phrase or even into one word.
 
The resulting list of groups resembles the list by~\citet{tarasevitch1991soviet} and looks as follows:

\begin{itemize}
    \item Lexical change
    \item Grammatical change
    \item Obsolescence
    \item Replacement
\end{itemize}

\begin{table*}[t]
\centering
\begin{tabular}{|l|l|l|}
\hline \bf MWE & \bf Criterion & \bf Annotation \\ \hline
\selectlanguage{russian} Белый гриб & \bf Lexical change & \\
(penny bun) & i. Lexemes (partially) lose independent meaning & 1 \\
& ii. Subordinate words cannot be replaced with a synonym & 1 \\
& iii. Headword cannot be replaced with a synonym & 1 \\
& iv. Cannot be translated word by word & 1 \\
& v. Does not allow insertions of lexemes & 1 \\
& \bf Grammatical change & \\
& vi. Never changes grammatical form & 0 \\
& vii. Only headword changes grammatical form & 0 \\
& viii. Word order seldom or never changes & 1 \\
& \bf Obsolescence & \\
& ix. Contains lexical archaisms & 0 \\
& x. Contains unique lexemes & 0 \\
& xi. Archaic syntax and/or morphology & 0 \\
& \bf Replacement & \\
& xii. Allows ellipsis* & 1 (can be called just \selectlanguage{russian} белый) \\
& xiii. Allows portmanteau words & 0** \\
& xiv. Can be replaced with headword & 1 \\
& xv. Can be replaced with one word & 1 (\selectlanguage{russian} белый, боровик)\\
& xvi. Can be translated with one word*** & 1 (cep, porcino) \\

\hline
\end{tabular}
\caption{An example of annotation of 16 criteria. *By ellipsis, we mean that a word or more can be omitted without change of meaning. **When words are ``sewn'' into one word. In this example, we disregard a very rare adjective \selectlanguage{russian}белогрибный\selectlanguage{british}. ***We checked translation into (sic!) English.}
\label{tab16props} 
\end{table*}

Note that in {\em Grammatical change} two criteria are mutually exclusive ({\em vi. Never changes grammatical form} and {\em vii. Only headword changes grammatical form}). Hence, an expression cannot score more than 15. Also, criteria iii., vii., xiv. are not applicable to MWEs with a sentence-like structure (containing a subject and/or predicate) and expressions with coordination due to absence of a headword. Criterion ii. is applied to all subordinate words in MWEs: if at least one of them can be replaced with a synonym, the score is 1. In cases when a criterion is not applicable it was annotated with 0.

To our annotation, we also added 4 linguistic features that are not directly bound to idiomaticity, but are often found as arbitrary criteria: POS-pattern, ``Is a sentence?'' (whether the MWE contains a subject and/or predicate), headword (if it is not a sentence), phrase structure (e.g. government, agreement, etc.). An example is given in Table~\ref{tab4props}. These features can be used in further research. All annotated data can be found at~\url{REDACTED FOR ANONYMITY}.

\begin{table*}[t]
\centering
\begin{tabular}{|l|l|l|l|l|}
\hline \bf MWE & \bf POS-pattern & \bf Is a sentence? & \bf headword & \bf Phrase structure \\ \hline
\selectlanguage{russian} Белый гриб 
\selectlanguage{british} (penny bun) & Adj.+Noun & No & \selectlanguage{russian} гриб & Agreement \\
\hline
\end{tabular}
\caption{An example of annotation of 4 linguistic features.}
\label{tab4props} 
\end{table*}

{\bf Gold standard of MWEs.} For annotation we took 286 examples of MWEs from the works by Russian scholars mentioned in~\ref{sec:theory}. The choice of the language was conditioned by our team being experts in the Russian language. However, we later describe how our results can be applied to other languages. The MWEs were collected without any pre-selection -- all examples that we could find in the papers excluding duplicates.

{\bf Expert annotation.} Our annotators, experts with higher education in linguistics, were instructed about the criteria described above. One expert annotated a column, then another expert looked it through and checked if they agree with the result. Questionable cases were discussed at seminars and led to a final decision about the annotation. However, we expect that the annotation can slightly change due to new arguments about this or that case~\footnote{It would be impossible to manually test some of the properties in all possible contexts, even using NLP-tools for support.}. Also, although the experts used grammar and other reference books, dictionaries, corpora and web to check their expertise, it is possible that they could overlook something, or still disagree even after the final decision about an MWE was made~\footnote{Many such cases were about the question whether a word is already obsolete.}. Hence, our annotation should be considered as an expert {\em approximation} of the real world.

To double-check some of the experts' annotation with NLP tools, we used Russian textual corpora~\footnote{https://ruscorpora.ru/}. The criterion ``v. Does not allow insertions of lexemes'' was checked with an expression ``WORD1 * WORD2'' where * shows that any word can be placed inside an MWE. The group of criteria {\bf Grammatical change} was checked similarly with * replacing the grammatical morphemes. In cases when only one corresponding example was found, we treated it as an occasional creative use of the language and, hence, equal to zero.

{\bf Clustering visualisation.} The aim of our visualisation is to find patterns that can provide us with a idea of how to cluster MWEs in the multi-dimensional vector space. To demonstrate it, we use a 3D scatter plot. Points are calculated as the vector sum in each {\bf group}. E.g. the vector for \selectlanguage{russian}``белый гриб''\selectlanguage{british} from Table~\ref{tab16props} is (5, 0, 0, 4). Limited by the three dimensions, to visualize the distribution of criteria, we take one group at a time and demonstrate with a color the average score of all MWEs that fall into this point. For example, only 1 MWE has the same score for the first three groups as 
\selectlanguage{russian}``белый гриб''\selectlanguage{british}: (5, 0, 0). Its fourth score is 1. The sum of the fourth scores is 4+1=5. Divided by the number of MWEs, it is now 2.5, and this number determines the color of the point.

\section{Data Analysis}
\begin{figure*}[t]
  \centering
  \includegraphics[width=1\linewidth]{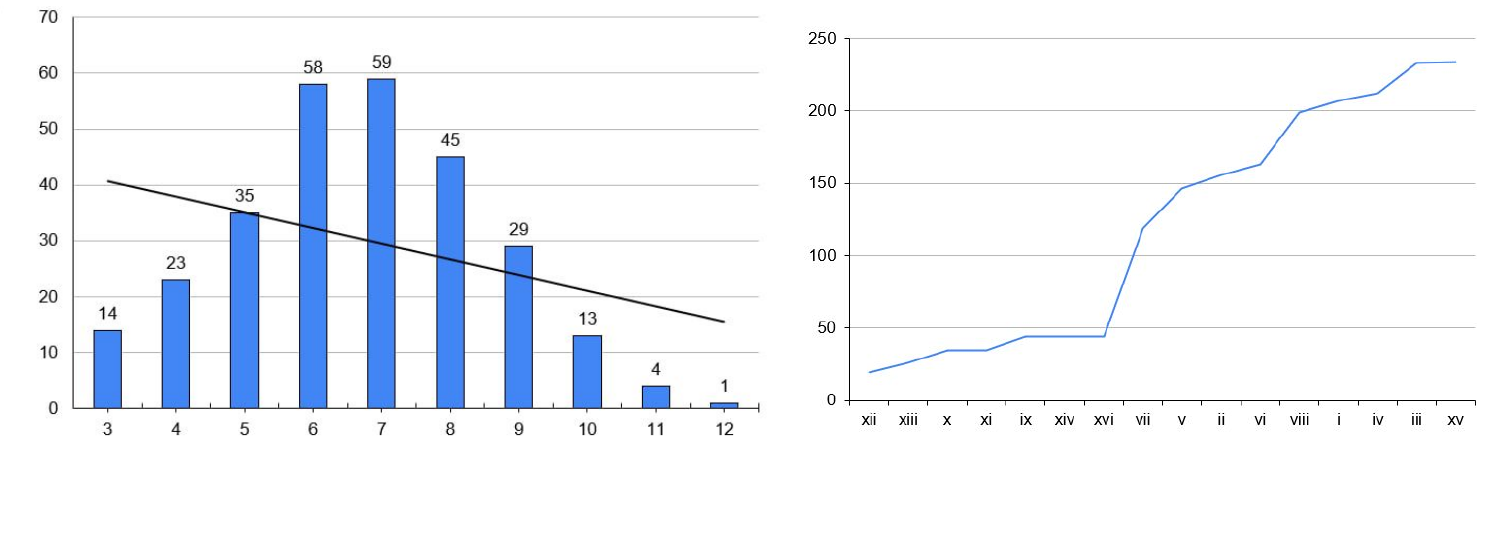}
  \caption{Number of MWEs scoring from 3 to 13 -- left. Sum of scores for each of the 16 categories (ranged) -- right.}
  \label{distribution}
\end{figure*}

We will now describe our resulting dataset and then attempt to find patterns in our 3D visualization. Due to our criteria being demonstrative of idiomaticity, we expect that the more of them score 1 for a given expression, the more idiomatic it is. Figure~\ref{distribution}, left, shows distribution of scores among the vectors. The distribution is right-skewed: 191 MWEs (67\%) score below the median -- on average, they do not tend to score much in all the groups of criteria immediately. And there are no MWEs that score from 13 to 15. Hence, the perfect MWE does not exist: some of the criteria either exclude others or make it very difficult to combine them in one expression.

The lowest idiomaticity score (vector sum) achieved is 3 -- for 14 MWEs. It is hard to notice any common features in them: \selectlanguage{russian} богу ведомо, всякой твари по паре, заклятый враг, игральные карты..\selectlanguage{british} ({\em God knows, motley crew, fast foe, playing cards..}). The highest score is 12: \selectlanguage{russian} так и быть\selectlanguage{british} ({\em So be it!}). And close to it are four expressions with the score of 11: \selectlanguage{russian}была не была; должно быть; ему и горюшка мало!; при пиковом интересе\selectlanguage{british} ({\em Come what may!; It must be that..; He doesn't give a damn!; (leave someone) holding the bag}). We note, that this set is more ``sentence-like'', tends to have a subject and a predicate.

The sum of scores (ranged from smallest to highest) for every criterion in Figure~\ref{distribution}, right, shows that several of them have very low values for the curve to be smooth. These are: {\em x. Contains unique lexemes; xi. Archaic syntax and/or morphology; ix. Contains lexical archaisms;  xiv. Can be replaced with headword; xvi. Can be translated with one word}. This can be due to a difficulty in annotating with these criteria. Translation requires a very good knowledge of foreign language (English, in our case); obsolete and unique words need vast expertise in both old and modern Russian; also, it can be hard to distinguish between an archaism and a bookish, formal modern word.

Some of the criteria score very high (have positive values for many MWEs -- Figure~\ref{distribution}, right, the right part of the curve), and that means for our task that they do not contribute to a stricter division between probable classes (they fill the vector space densely). The top-scoring (above 200) criteria are: {\em i. Lexemes (partially) lose independent meaning (207 MWEs), iv. Cannot be translated word by word (212 MWEs), iii. Headword cannot be replaced with a synonym (233 MWEs), xv. Can be replaced with one word (234 MWEs)}. Intuitively, Criterion i. appears like it is what theorists look for in all idiomatic expressions, regardless of their type. Also, these criteria seem to be based more on lexical rather than grammatical properties.

\subsection{Patterns in Distribution of Vectors}

Out of 286 MWEs 216 are unique vectors (76\%). It is hard to say whether our annotation exhibits strong general patterns in distribution of all the criteria across the dataset. We will try to generalize about what happens in each of the defined groups.

{\bf Lexical change.} This group is the champion in earned scores: 954 (50\% of all scores). Which, probably, means that it embodies idiomaticity and should not be considered as a basis of ``MWEhood''. I.e. any candidate for an MWE should score high in it. An example of a low-scorer in this group is \selectlanguage{russian}беспокойный человек\selectlanguage{british} {\em a restless person}. This is what we earlier called an MWU, an expression that learners of a language memorize to sound more like native speakers, and it scores only 4 in all categories. We believe that {\em Lexical change} should be used for classifying between MWEs and non-MWEs and, probably, for defining MWUs.


{\bf Grammatical change.} This category has a very high score (481, 25\%) and can be called influential as well. Although there can be the mentioned bias in selection of examples: those selected by scholars tend to be more fixed structurally, to better demonstrate fixedness. Expressions that score 0 in it are, for example, (\selectlanguage{russian}беспробудное пьянство, потрясающее впечатление\selectlanguage{british} {\em deep drinking, stunning impression}) -- also typical MWUs. Also, presence of a verb usually makes expressions in Russian (as well as in English) less fixed. Compare:\selectlanguage{russian} пуститься во все тяжкие -- во все тяжкие\selectlanguage{british} {\em do whatever it takes -- whatever it takes}; the verb {\em do} can change its grammatical form and can be substituted with a synonym. This leads us to one of the main conclusions that what is singled out as an MWE can be be several expressions of a different degree of idiomaticity.

{\bf Obsolescence.} As we mentioned earlier, this category seldom earned a positive decision from our annotators. There are only 9 collocations that score 3 in it (e.g.\selectlanguage{russian} и вся недолга!; не до жиру, быть бы живу; ничтоже сумняшеся\selectlanguage{british} {\em end of story, survive before thrive; nothing doubting}). The collocations from our example score 8 and more in all criteria (with 7.5 being the median value, and 7.4 -- mean). Although it is probably clear without the data, but our experiment supports it that this category is a strong marker of idiomaticity. To add, obsolete words make it harder to modify an expression; archaic grammar hinders morphological and other changes in different contexts.

{\bf Replacement.} Counter-intuitively, this category seems to be in a conflict with {\em Grammatical change} and {\em Obsolescence}. Often, if an MWE scores high in it, it scores low (or medium) in one or both. The example in Table~\ref{tab16props} demonstrates it. There are 231 expressions that score less than 3 both in {\em Obsolescence} and {\em Replacement} which might mean that these two categories are not crucially important in formation of an WME, but they probably point at two distinct sub-classes.

\begin{figure*}[t]
  \centering
  \includegraphics[width=0.9\linewidth]{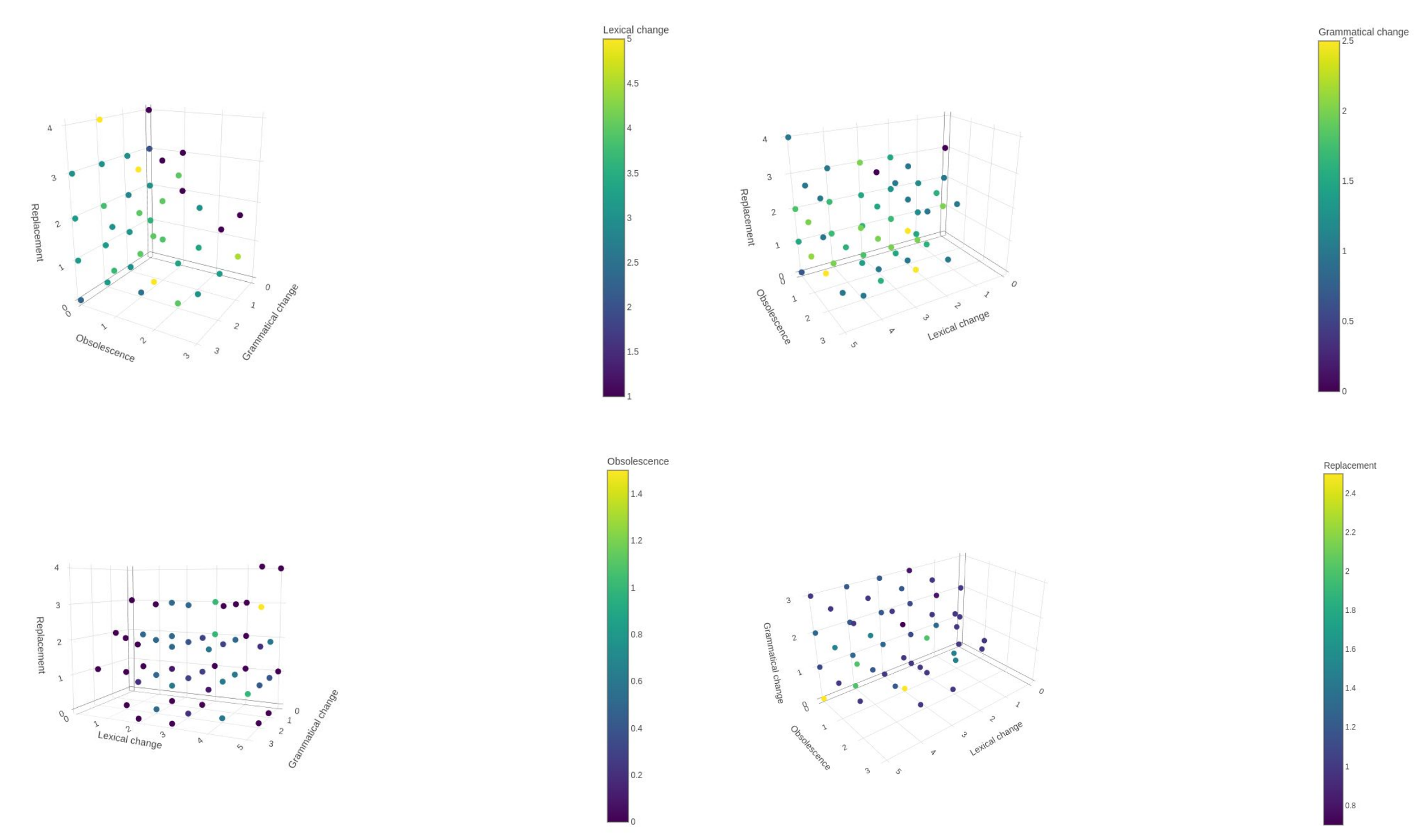}
  \caption{3D scatter plot of the sums of scores of MWEs in each category.}
  \label{3d_diagram}
\end{figure*}

{\bf 3D-model of vector distributions.} We now want to demonstrate how the vectors are distributed in a multi-dimensional space with a 3D scatter plot. We described in Chapter~\ref{sec:experset} how the cubes were built. In each of the four cubes in Diagram~\ref{3d_diagram}, color demonstrates the average value of the fourth category. The angle is chosen so as to illustrate our hypotheses.

The top-left cube where the color points the average value of {\em Lexical change} shows that there is little correlation between it and {\em Replacement}: with any value of {\em Replacement} {\em Lexical change} can be high (yellow and green points). It positively correlates with {\em Grammatical change} and {\em Obsolescence}.

It looks like {\em Grammatical change} (top-right) requires certain conditions to score high: [2:4] for {\em Lexical change} and [1:2] for {\em Replacement}. 

{\em Obsolescence} (bottom-left) positively correlates with {\em Lexical change}, but it is unclear how it correlates with the two other groups.

And the group {\em Replacement}, bottom-right, also positively correlates with {\em Lexical change} and negatively correlates with {\em Obsolescence}. It does not seem to correlate with {\em Grammatical change}.

\section{Conclusion}

The paper describes an attempt to search for theoretical grounds in the notion of idiomaticity with the help of linguistic annotation and data analysis of the gold standard MWEs. We have proposed a model of 16 criteria that were grouped into four categories based on linguistic analysis. We annotated a corpus of 286 Russian MWEs found in the same theoretical books from which we took the criteria. Our analysis revealed several trends:

\begin{itemize}
    \item the group of criteria that we called {\em Lexical change} positively correlates with other criteria which makes it look rather like a target category and an idiomaticity ``test''. A better proof requires comparison to non-MWEs;
    \item either scholars tend to choose MWEs that show medium {\em Grammatical change} or it is a stable property of all such expressions;
    \item some MWE criteria are, probably, harder to annotate (e.g. identifying obsolete words and contexts allowing to replace an MWE);
    \item there is no or negative correlation between presence of archaic words and grammar and the property of an expression to be shortened or replaced by a single word;
    \item what is determined as an MWE can be several expressions with a different degree of idiomaticity.
\end{itemize}

{\bf Relation to other languages.} It is hard to say whether the same conclusions will stand for other languages. As we mentioned, some of the described features can be already observed in English, e.g. verbs in MWEs make expressions less idiomatic. Our annotated dataset contains some translation into English, and also examples can be taken from the studied English literature to create a similar corpus. A further extension can lie in application of NLP-tools to automatic annotation of such corpora as PARSEME collections with the criteria that we described. 

{\bf Future work}. Future work requires a larger annotated collection. The main obstacle here is that manual annotation that we performed is very time-consuming. We can see several more ways, beside the mentioned ones, of making it partially automatic, e.g. impossibility of translating an MWE word by word as well as translation with one word can be checked in parallel corpora. Also, the criteria should be checked in free word groups. And finally the stated correlations require quantitative analysis.

\bibliographystyle{lrec_natbib}
\bibliography{dialogue,custom}

\end{document}